# INTELLIGENT PAINTER: PICTURE COMPOSITION WITH RESAMPLING DIFFUSION MODEL

*Wing-Fung Ku[1], Wan-Chi Siu[1,2] (Life-FIEEE), Xi Cheng[1] and H. Anthony Chan[1] (FIEEE)*
[1]School of Computing and Information Sciences, Caritas Institute of Higher Education
[2]Department of Electronic and Information Engineering, The Hong Kong Polytechnic University

## ABSTRACT

Have you ever thought that you can be an intelligent painter? This means that you can paint a picture with a few expected objects in mind, or with a desirable scene. This is different from normal inpainting approaches for which the location of specific objects cannot be determined. In this paper, we present an intelligent painter that generate a person's imaginary scene in one go, given explicit hints. We propose a resampling strategy for Denoising Diffusion Probabilistic Model (DDPM) to intelligently compose unconditional harmonized pictures according to the input subjects at specific locations. By exploiting the diffusion property, we resample efficiently to produce realistic pictures. Experimental results show that our resampling method favors the semantic meaning of the generated output efficiently and generates less blurry output. Quantitative analysis of image quality assessment shows that our method produces higher perceptual quality images compared with the state-of-the-art methods.

## 1. INTRODUCTION

Our minds are good at fabricating scenes. However, it is difficult to show our pictured mind to other people with words, which is implicit. Some people could draw the picture from their minds, but drawing is a skill that not everyone has expertise. It is always easy to imagine a single image object. One or a few objects can form the landmark information of a picture. For example, if we know a scene to have a house and a tree next to the house, we can imagine that the ground is grassland, and the sky is located above the house. We, therefore, can compose a picture when explicit hints are given.

For scene composition, it involves scene generation. This makes us to recall different Generative Adversarial Networks (GANs) [1] methods, including iGAN [2], GANBrush [3], and PoE-GAN [4]. However, the above methods make uses of image priors which limit the user inputs into certain types. In this paper, we work on an image completion problem, since we are filling the unknown pixels according to our known components which are the landmarks. Early image completion approaches either rely on a large image database for matching image regions [5] or make use of the neighbor pixels to approximate the missing region [6]. They were only effective when repairing small patches. However, in later approaches researchers started to use deep learning methods [7-13] which have the generalization ability to support large incompletion area.

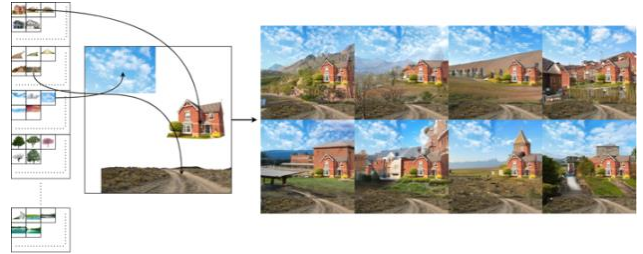

**Fig. 1.** Proposed objective of our intelligent painter

The following gives contributions of this paper. 1) We initially suggest that every reader can be an intelligent Painter for compositing an ideal picture with key items from his/her mind, which should be a big future direction in deep learning applications. 2) We propose a resampling strategy to fill out large empty space to form high-quality pictures with unconditional DDPM. 3) Our inference time is much faster, with better picture appearance and quality metric scores when comparing with RePaint [7] which also uses DDPM.

## 2. PROBLEM FORMULATION AND BACKGROUND CONCEPTS

Basically, our model is able to take user's input items as keep components and guide the missing area such that the filled area seamlessly matches with the input. This is done by encoding the known input information $x^{known}$ into latent $z^{known}$ and combine it with the existing trained latent $z^{unknown}$. Then by decoding the merged latents, we are able to obtain the seamless picture y.

$$z^{known} = Encode(x^{known}) \qquad (1)$$
$$y = Decode(z^{known} + z^{unknown}) \qquad (2)$$

In order to achieve our formulation, we need a model for which the latent has the same dimension with our output to allow combination. In this case diffusion model may be a possible choice as its latent variable has the same dimension as the data and output.

Denoising Diffusion Probabilistic Model (DDPM) is a generative model that could perform high-quality image synthesis [16]. It only learns the decode stage. To be specific, it denoises an image in a scheduled manner, starting from a pure noise image which is similar to latent. The setup of DDPM consists of a fixed forward process $q$ which holds similar idea to encoding, and a learned reverse process $p_\theta$ to decode. Both processes are indexed by $t$ in the finite timesteps $T$ and formed a Markov chain. In each step of the fixed forward process $q$, a different scale of noise according to the timestep is applied to the input image $x_0$. The noise accumulates for each timestep until the image forms complete random noise image $x_T$, which can be viewed as an encoding process. A step of the forward process is a conditional probability defined as a normal distribution function $\mathcal{N}(x; \mu, \sigma^2)$.

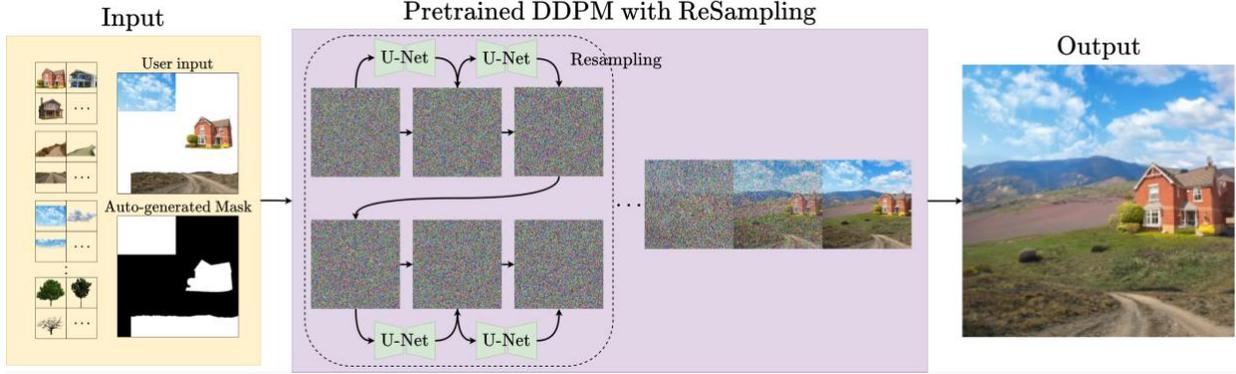

**Fig. 2** Architecture of proposed Intelligent Painting

$$q(x_t \mid x_{t-1}) = \mathcal{N}(x_t; \sqrt{1-\beta_t} x_{t-1}, \beta_t \mathbf{I}) \quad (3)$$

Given image $x_{t-1}$, we can find the probability of image $x_t$. Schedule $\beta$ is a $T$-shaped vector with interval from 0.0001 to 0.02. Since normal distributions with any mean and variance can be represented as a scaled and translated standard normal distribution $\mathcal{N}(0,1)$, we could use reparameterization trick [17] to convert $\mathcal{N}(\mu, \sigma^2) \to \mu + \sigma \cdot \epsilon$, where $\epsilon$ is the noise sampled from $\mathcal{N}(0, \mathbf{I})$ and $\mathbf{I}$ is an identity matrix. Thus, the formula can be simplified as:

$$x_t = \sqrt{1-\beta_t} x_{t-1} + \sqrt{\beta_t} \epsilon \quad (4)$$

To efficiently apply accumulated noise to a specific time step from the initial image, we can make use of the gaussian distribution property and calculate the total noise variance. We can derive the formula for apply any number of forward steps as one step:

$$\sqrt{\alpha_t \alpha_{t-1} \dots \alpha_1 \alpha_0} x_0 + \sqrt{1 - \alpha_t \alpha_{t-1} \dots \alpha_1 \alpha_0}\, \overline{\epsilon}$$
$$= \sqrt{\overline{\alpha_t}} x_0 + \sqrt{1-\overline{\alpha_t}}\, \overline{\epsilon} \quad (5)$$

where $\overline{\alpha_t} = \prod_{s=1}^{t} \alpha_s$. We note that a merged gaussian $\overline{\epsilon}$ can be represented as a gaussian $\epsilon$. Therefore, the equations of the forward process can be simplified as:

$$q(x_t \mid x_0) = \mathcal{N}(x_t; \sqrt{\overline{\alpha_t}} x_0, (1-\overline{\alpha_t}) \mathbf{I}) \quad (6)$$
$$x_t = \sqrt{\overline{\alpha_t}} x_0 + \sqrt{(1-\overline{\alpha_t})} \epsilon \quad (7)$$

For the learned reverse process $p_\theta$, a U-Net [18] with learnable parameter set $\theta$ and the time embedding [19] is trained to predict the noise added at specific timestep $t$, such that given image $x_t$, image $x_{t-1}$ can be derived by removing one step of noise. To generate meaningful content for our intelligent painter, we can sample a pure noise image $x_T$ and chain all the reverse processes until $x_T$ become a semantically meaningful image $x_0$. A step of the reverse process is also a conditional probability which can be formulated as a parameterized distribution function. This can be simplified with reparameterization trick as well.

$$p_\theta(x_{t-1} \mid x_t) = \mathcal{N}(x_{t-1}; \mu_\theta(x_t, t), \Sigma_\theta(x_t, t)) \quad (8)$$
$$x_{t-1} = \mu_\theta(x_t, t) + \sqrt{\Sigma_\theta(x_t, t)}\, \epsilon \quad (9)$$

where the mean $\mu_\theta(x_t, t)$ and variance $\Sigma_\theta(x_t, t)$ are the output of the trained U-Net.

## 3. PROPOSED RESAMPLING ON DDPM FOR SCENE COMPOSITION

To make scene composition possible, instead of training a DDPM in a conditional input setting, we modify the inference process of an unconditional pre-trained DDPM to achieve our result. We set our landmark input to guide the denoising process. In details, for each reverse step, we encode our image $x_0$ to the noisy image $x_{t-1}^{known}$, while decodes a random noise image $x_t$ to obtain $x_{t-1}^{unknown}$. The two intermediate results are masked to keep the landmark region and then added together before the next step. In such a way, the landmark information encoded could guide the content generation in the unknown area. It can be formulated as the following equations:

$$x_{t-1}^{known} = \sqrt{\overline{\alpha_t}} x_0 + \sqrt{(1-\overline{\alpha_t})}\, \epsilon \quad (10)$$
$$x_{t-1}^{unknown} = \mu_\theta(x_t, t) + \sqrt{\Sigma_\theta(x_t, t)}\, \epsilon \quad (11)$$
$$x_{t-1} = m \odot x_{t-1}^{known} + (1-m) \odot x_{t-1}^{unknown} \quad (12)$$

where m is the mask generated from our input as shown in Fig. 2, $\odot$ means element-wise product. The equations are essentially the same as our problem definition in equation (1) and (2). However, the above algorithm only considers the conditional landmark information and tries to fill in the blank parts. The algorithm fails to maintain the perceptual quality of the output image for our task as it fails to incorporate the semantic information, especially when the conditional landmark information is insufficient.

Therefore, let us add a resampling to help the DDPM to consider the generated parts of the image, analogous to those proposed in RePaint [7]. Resampling is done by reintroducing the forward process that the less noisy image $x_{t-\lambda}$ is encoded back to a noisier image $x_t$, and then denoised back to image $x_{t-\lambda}$ again. In that case, the generated information $x_{t-\lambda}$ is preserved in $x_t$, resulting in a more semantically meaningful output. We have found that to make the resampling effective for our application, we can set λ as 10 which means we apply 10 steps of the forward process before 10 denoising steps. The resampling is repeated 10 times for each of the 10 denoising timesteps across the schedule, as Lugmayr et. al. [7] suggested that the higher number of resampling would improve the overall image consistency.

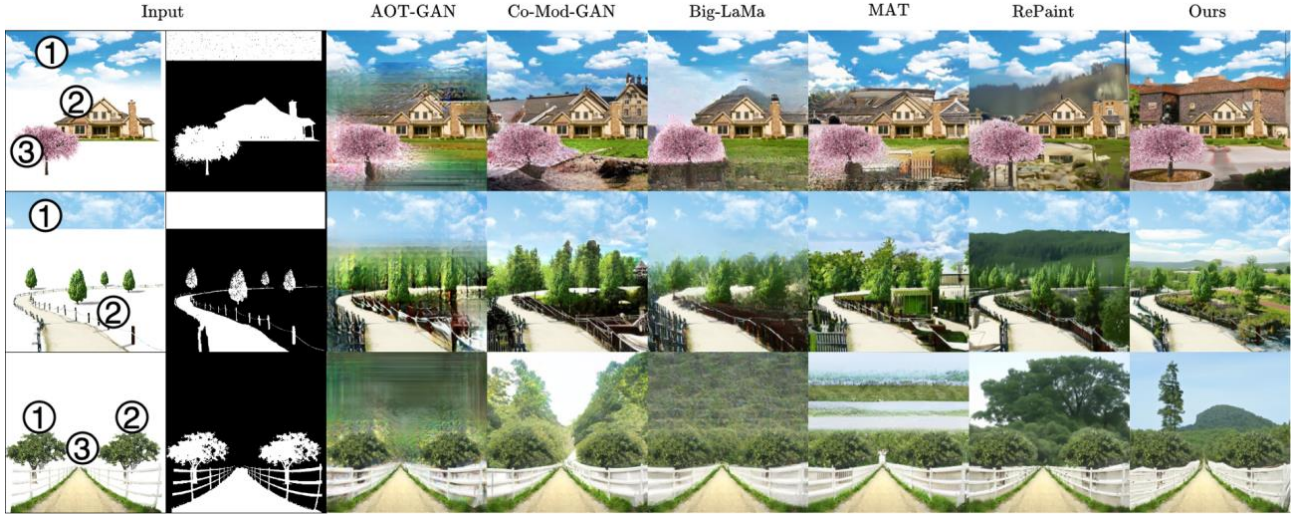

**Fig. 3** Visual comparison with different methods

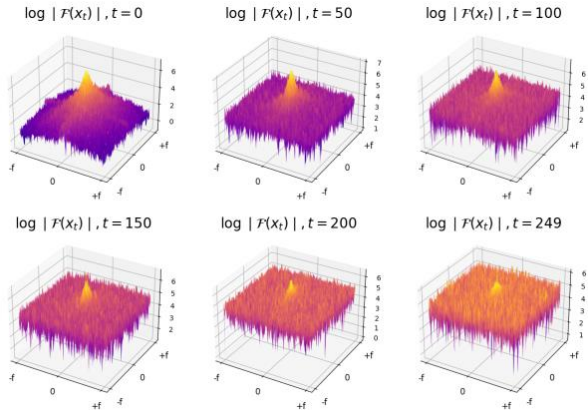

**Fig. 4** Fourier visualization of DDPM forward process

One serious downside is that resampling increases the inference time significantly, as it increases the operations. Moreover, the resampling approach proposed by RePaint [7] often produces image with unclear details. To mitigate this issue, we stop the resampling at timestep $t = 100$. According to the property of DDPM fixed forward process $x_t = \sqrt{\bar{\alpha}_t}x_0 + \sqrt{(1-\bar{\alpha}_t)}\epsilon$, when the timestep t is small, $\bar{\alpha}$ is a value close to 1. Alternatively, $\bar{\alpha}$ is close to 0 when the timestep is big. Using the linear property of the Fourier transform, we perform Fourier analysis on the forward process $\mathcal{F}(x_t) = \sqrt{\bar{\alpha}_t}\mathcal{F}(x_0) + \sqrt{(1-\bar{\alpha}_t)}\mathcal{F}(\epsilon)$ and found that the high-frequency components of the image such as fine details are corrupted at lower timesteps, while at larger timesteps the low-frequency components of the image such as coarse structures are corrupted, as shown in Fig. 4. Therefore, we can assume that the learned reverse process first generates the coarse structure at higher timesteps and then makes fine details at lower timesteps. The resampling at lower timesteps could potentially blur the details because the preserved low-level information in the forward steps has intervened the denoising process at lower timestep. Since the coarse structure harmonized at bigger timesteps has provided enough information to generate the fine details, we can exploit this property to stop resampling at a smaller timestep, therefore improving the image perceptual quality and the inference time.

## 4. EXPERIMENTAL RESULTS AND DISCUSSION

### 4.1 Visual comparison with other methods

We compared our approach with some state-of-the-art models including Big-LaMa [8], AOT-GAN [9], MAT [10], Co-Mod-GAN [11], and RePaint [7]. All models were trained on the Place2 dataset to ensure a fair comparison. We found that GAN-based models (AOT-GAN, Co-Mod-GAN) often produce blurry artifacts, and Fourier-based model (Big-LaMa) produces repetitive artifacts. Transformer-based model (MAT) produces better results, but sometimes failed to consider the semantic meaning. Our diffusion-based models result in higher perceptual quality as illustrated in Fig. 3.

### 4.2 Quantitative metric comparison

Validating the effectiveness of this approach is challenging because our task has no ground truth. Common image quality assessment metrics such as SSIM [20] and PSNR are not usable. To quantitatively validate our approach, we make use of NR-IQA (No-Reference Image Quality Assessment) methods [21-24] to access the quality of the output image.

TABLE I. Quantitative evaluation

| Method | FID [22](↓) | NIQE [21](↓) | HyperIQA [23](↑) | NIMA [24](↑) |
|---|---|---|---|---|
| AOT-GAN[9] | 277.83 | 5.96 | 44.75 | 5.22 |
| CoModGAN[11] | 225.62 | 5.30 | 46.42 | 5.28 |
| Big-LaMa[8] | 243.55 | 4.95 | 45.41 | 5.10 |
| MAT[10] | 231.15 | 5.65 | 44.66 | 5.24 |
| RePaint[7] | 238.88 | 5.56 | **47.86** | **5.38** |
| Ours | **221.20** | **4.86** | 47.64 | 5.33 |

From Table I, we can see that our method achieves the lowest FID [21] and NIQE [22], which indicates the best image quality.

### 4.3 Ablation study on resampling strategies

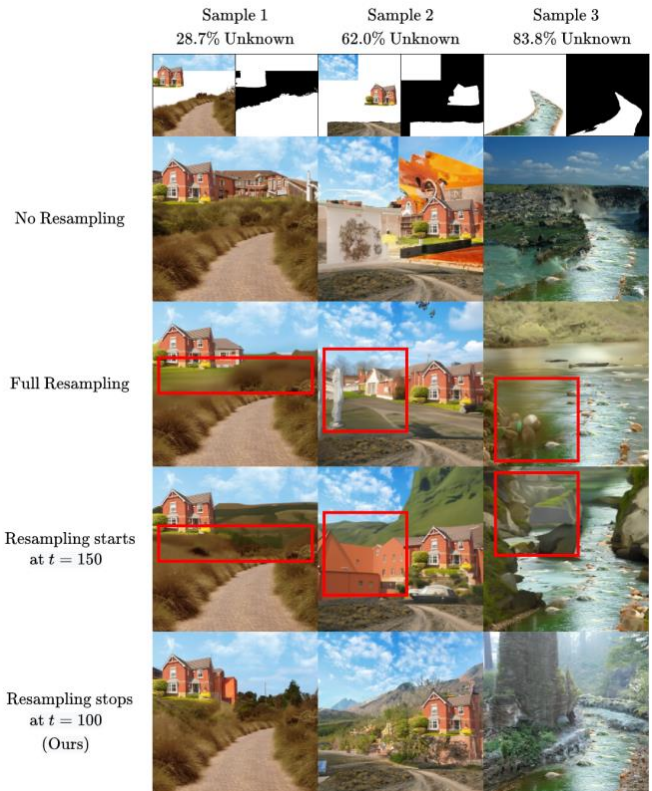

**Fig. 5** Visual comparison on resampling strategies

In Fig. 5, we compare the output images among various resampling methods. We can find that without resampling the outputs are inconsistent and lack of semantic meaning, especially when the unknown area is large. With the resampling technique, the output is more harmonized and consistent. However, we also found that for full resampling setting the results suffer from unclear details problem. We found the unclear details problem is similar to the stochastic variation technique problem in StyleGANv1 [25]. We have identified the problem and found that performing resampling at lower timesteps tends to yield blurry images. These blurry artifacts appeared frequently on the approach "resampling starts as t = 150" and "full sampling". With this observation, we stop resampling at t = 100 to harmonize the coarse structures while keeping the detailed information untouched. Our resampling method not only shortens the inference time but also improves the perceptual quality of the output images. To further analyze the number of operations, we note that multiple steps of the forward process can be done in one step. From Table II, we can see that even though the accumulated forward process can be done in 1 step, the operating steps increase linearly. The inference time comparison is shown on Table III. Diffusion models are generally slow due to the time of sampling but provide better diversity. Our method is 40% faster than RePaint [7] while achieving better perceptual quality in our task as shown in Table I.

TABLE II. number of operations of resampling strategies

| | $N^{dn}$ | $N^{fwd}$ | $N^{total}$ |
|---|---|---|---|
| Resampling All | 2410 | 216 | 2626 |
| Start resampling at T = 150 | 1600 | 135 | 1735 |
| Stop resampling at T = 100 | 1510 | 126 | 1636 |
| No Resampling | 250 | 0 | 250 |

TABLE III. Inference time comparison on a RTX 3090 GPU

| Method | Inference time in seconds (↓) |
|---|---|
| AOT-GAN[9] | 0.4961 |
| CoModGAN[11] | 0.5331 |
| Big-LaMa[8] | **0.0353** |
| MAT[10] | 0.0694 |
| RePaint[7] | 321.6836 |
| Ours | **201.8138** |

## 5. CONCLUSIONS

Controllable image synthesis is a hot topic nowadays, and researchers have designed novel approaches to generate high-quality image components, allowing users to produce their desire images. However, researchers ignored the possibility of fabricating images by injecting real-life images. In this paper, we propose the use of explicit landmarks and a resampling strategy to perform content-guided image synthesis using DDPM. Our resampling strategy significantly reduced the inference time while perceiving the perceptual quality of the image. In details, we exploit the frequency property of DDPM and force the model to only resample the low-level content, efficiently yielding the outputs that are semantically harmonized. Our method also tends to produce a less blurry result as compared with the full resampling strategy by RePaint and has a faster inference time. Compared with the state-of-the-art approaches, our proposed method achieves better NIQE and FID scores which imply better image quality. To further improve this work and provide more controllability, we will research the multimodal intelligent painter. Program codes are available at github.com/vinesmsuic/ipainter-diffusion.

## 6. ACKNOWLEDGEMENTS

This work is partly supported by the Caritas Institute of Higher Education (ISG200206) and UGC Grant (UGC/IDS(C)11/E01/20) of the Hong Kong Special Administrative Region.